\documentclass[10pt,twocolumn,letterpaper]{article}

\usepackage{3dv}
\usepackage{times}
\usepackage{epsfig}

\usepackage[english]{babel}
\usepackage[utf8]{inputenc}
\usepackage{xargs} 
\usepackage[pdftex,dvipsnames]{xcolor}  
\usepackage{graphicx} 
\usepackage{amssymb}
\usepackage{amsmath}
\usepackage{mathtools} 
\usepackage[super]{nth} 
\usepackage{physics} 
\usepackage{enumitem} 
\usepackage{gensymb} 
\usepackage{booktabs} 
\usepackage{tabularx} 
\usepackage{caption} 
\usepackage{subcaption} 
\usepackage{cite} 

\makeatletter
\@namedef{ver@everyshi.sty}{}
\makeatother
\usepackage{tikz} 
\usepackage{tkz-euclide} 

\usepackage{pgfplots} 
\usepackage[colorinlistoftodos,prependcaption,textsize=tiny]{todonotes}
\usepackage{csquotes} 
\usepackage{xr-hyper}  
\usepackage[pagebackref=true,breaklinks=true,letterpaper=true,colorlinks,bookmarks=false]{hyperref}
\usepackage{cleveref}

\MakeOuterQuote{"}

\usetikzlibrary{positioning, arrows}

\addto\extrasenglish{

}

\makeatletter
\let\oldtheequation\theequation
\renewcommand\tagform@[1]{\maketag@@@{\ignorespaces#1\unskip\@@italiccorr}}
\renewcommand\theequation{(\oldtheequation)}
\makeatother

\creflabelformat{equation}{#2\textup{#1}#3}
\renewcommand{\autoref}{\Cref}

\newcommandx{\Todo}[2][1=]{\todo[linecolor=yellow,backgroundcolor=yellow!25,bordercolor=yellow,#1]{#2}}
\newcommandx{\info}[2][1=]{\todo[linecolor=green,backgroundcolor=green!25,bordercolor=green,#1]{#2}}

\newcolumntype{Y}{>{\centering\arraybackslash}X}

\allowdisplaybreaks 

\title{Error Bounds of Projection Models in Weakly Supervised 3D Human Pose Estimation}

\author{
  Nikolas Klug\textsuperscript{1*} \;\;\; Moritz Einfalt\textsuperscript{2*} \;\;\; Stephan Brehm\textsuperscript{2} \;\;\; Rainer Lienhart\textsuperscript{2}\\
  University of Augsburg\\
  {\small\textsuperscript{1}\tt klug.nikolas@gmail.com} \\
  {\small\textsuperscript{2}\tt \{firstname.lastname\}@uni-a.de}
}

\threedvfinalcopy 

\def\threedvPaperID{214} 

\ifthreedvfinal\fi

\begin{document}
\maketitle

\begingroup\renewcommand\thefootnote{$*$}
\footnotetext{Authors contributed equally}
\endgroup

\thispagestyle{empty}

\begin{abstract}

  The current state-of-the-art in monocular 3D human pose estimation is heavily influenced by weakly supervised methods.
  These allow 2D labels to be used to learn effective 3D human pose recovery either directly from images or via 2D-to-3D pose uplifting.
  In this paper we present a detailed analysis of the most commonly used simplified projection models, which relate the estimated 3D pose representation to 2D labels: normalized perspective and weak perspective projections.
  Specifically, we derive theoretical lower bound errors for those projection models under the commonly used mean per-joint position error (MPJPE).
  Additionally, we show how the normalized perspective projection can be replaced to avoid this guaranteed minimal error.
  We evaluate the derived lower bounds on the most commonly used 3D human pose estimation benchmark datasets.
  Our results show that both projection models lead to an inherent minimal error between $19.3$mm and $54.7$mm, even after alignment in position and scale. This is a considerable share when comparing with recent state-of-the-art results.
  Our paper thus establishes a theoretical baseline that shows the importance of suitable projection models in weakly supervised 3D human pose estimation.
  
\end{abstract}

\section{Introduction}

\begin{figure}[t]
  \centering
  \includegraphics[width=0.8\linewidth]
                   {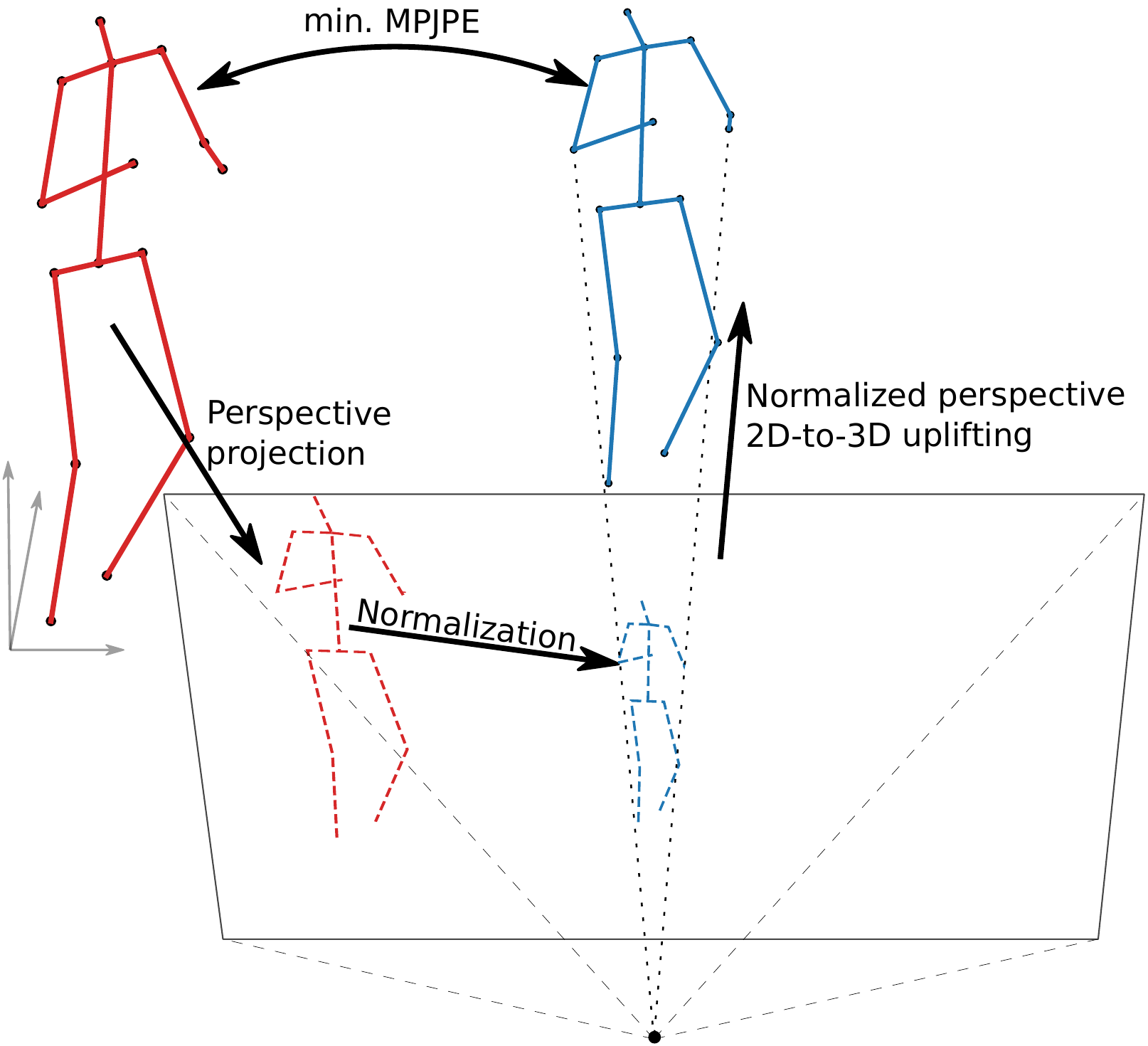}
  \caption{Lower bound error (MPJPE) for normalized perspective projection: We derive the best-case 2D-to-3D re-projection after 2D pose normalization and compare it to the original 3D human pose via MPJPE metric.
  }
  \label{fig:normalized_perspective}
\end{figure}

The improvements in 2D human pose estimation (HPE) over the last years have been one of the most prominent successes of deep learning in computer vision tasks.
Consequently, there is a still growing ambition to transition those advancements into the next logical step in human pose estimation: the monocular reconstruction of the 3D human pose, specifically from single images.
Successful architectures and representations from the 2D task have been re-used to build end-to-end 3D pose reconstruction models with convincing results \cite{pavlakos17, pavlakos18, nibali2019}.
But the transfer into 3D space comes with additional difficulties that need to be addressed.
First, labeled 3D human pose datasets are less available than large-scale 2D datasets.
They are much more limited in scale as well as variability of appearance and poses.
Second, the task of monocular 3D human pose reconstruction is less constrained and more difficult to optimize compared to its 2D counterpart \cite{popa17}.
Most prominently, this comes from the inherent ambiguities of object scale, depth and camera intrinsics.
Different combinations can lead to the same 2D projection.
This makes the reconstruction of the single, unique 3D pose impossible, unless strong priors about the object, scene and camera are given.

One approach to tackle both challenges can be seen in recent weakly supervised 3D HPE methods.
Typically, they use two-stage pipelines, where initially the 2D human pose is inferred from the image using existing methods.
Only this sparse 2D representation is fed into a separate deep neural network, which reconstructs the pose in 3D space.
The 2D-to-3D uplifting employs a generative approach, where the estimated 3D pose has to match a learned distribution of realistic human poses \cite{tung17, drover18, wandt19}.
It avoids the need for direct supervision.
One necessity is an explicit projection model to ensure a correct 3D-to-2D relationship.

Due to the ambiguities of monocular 3D reconstruction, current 2D-to-3D uplifting adopts simplified projection models. These add additional constraints that reduce the degrees of freedom of standard perspective projection with a pin-hole camera model.
The two most common choices are the perspective projection after 2D normalization and the \emph{weak perspective projection}.
The former, which we refer to as the \emph{normalized perspective projection}, simulates the perspective projection of a camera centered on the target human with a fixed projected scale.
It enforces any input 2D pose to be artificially normalized in position and scale before 2D-to-3D uplifting\cite{drover18}.
The weak perspective projection model approximates the true perspective projection with a scaled orthogonal projection \cite{kanazawa18, wandt19}.
Both variants do not match the original projection process in all but the most trivial cases.
This mismatch negatively affects the fidelity of the 3D pose estimate.

In this paper, we provide a detailed analysis of those simplified projection models.
Our goal is to derive a minimal error that these projection models are guaranteed to induce (\autoref{fig:normalized_perspective}).
For this, we develop a general formulation for the best-case 3D estimate that a 2D-to-3D uplifting network can find under those projections.
By evaluating the best-case estimates on popular 3D HPE benchmark datasets, we provide a quantitative lower error bound for weakly supervised 2D-to-3D uplifting networks.
It provides a baseline for the comparison of current fully and weakly supervised 3D human pose estimators.
Additionally, we present a relaxed version of the normalized perspective projection model that avoids the necessity for position normalization.
We integrate it into a baseline architecture and show how it can resolve the otherwise considerable error bound.
Our contributions can be summarized as follows:
\begin{itemize}
\item We derive theoretical lower bound errors for weakly supervised 3D HPE under the constraints of normalized and weak perspective projection models.
\item We evaluate the error bounds in synthetic experiments and on real-world 3D HPE benchmark datasets to identify the key factors that influence the minimal error.
\item We propose a relaxed projection model for weakly supervised pipelines that can improve upon the original error bound by a large margin.
\end{itemize}

\section{Related Work}

Recent literature on deep learning based monocular 3D HPE is vast, with various paradigms, representations and architectures.
We briefly cover recent work based on supervision type and the included projection models, if any, and relate it to our study.
Note that our work focuses on 3D HPE in single images. There is highly active research on recovering temporal 3D pose trajectories from monocular videos (e.g. \cite{zhou16, grinciunaite16, pavllo19, kanazawa19, kocabas2020}), which we do not consider here.

\textbf{Full supervision:} Fully supervised approaches for translating images to 3D human poses are conceptually the most similar to state-of-the-art 2D HPE models.
They usually build around fully convolutional architectures and spatially encoded regression targets \cite{pavlakos17, sun18, nibali2019, fabbri2020}.
Limited by the need for sufficient 3D-annotated images, multiple proposals have been made to simplify the overall objective or relax the need for perfect 3D annotations.
Kocabas \etal~\cite{kocabas2019} use noisy 3D poses from multi-view epipolar geometry as the training input for their 3D HPE CNN. Rogez \etal~\cite{rogez16} as well as Chen \etal~\cite{chen17} formulate 3D HPE as a selection task, where the best-fitting pose from a set of template poses is chosen.
Apart from direct image translation, Martinez \etal~\cite{martinez17} introduce a 2D-to-3D pose uplifting model with a fully supervised MLP.
It effectively decouples the variability in human appearance from the much more constrained space of possible human poses.
This paradigm has been prevalent in subsequently developed weakly supervised approaches.
Overall, those fully supervised methods do not require manually projecting 3D pose estimates back into the 2D input domain, alleviating the requirement for an explicit projection model.
In general, their performance is therefore not directly related to our established error bounds.

\textbf{Weak and mixed supervision:} Weakly supervised approaches for 3D HPE are usually characterized by 2D-to-3D pose uplifting in a generative framework.
This is achieved with GAN-like networks which simultaneously learn the 3D pose reconstruction as well as its distribution.
They enforce the 3D estimate to match the 2D input pose by integrating an explicit projection model.
The 2D-to-3D uplifting process essentially re-projects the 2D input into 3D space according to the inverse projection model, while maintaining a realistic 3D estimate.
Drover \etal~\cite{drover18} lift the normalized 2D input pose back into 3D space with respect to a depth estimate and a fixed perspective projection.
Additionally, random projections of the resulting 3D estimate have to match a learned distribution of realistic 2D human poses. Since the normalized perspective projection is a structural component, it acts as a hard constraint and makes it directly applicable to our derived error bounds.
Wandt \etal~\cite{wandt19} propose a similar model, but instead directly learn a distribution of realistic 3D human poses in a kinematic representation. The consistency between the 3D estimate and the 2D input is ensured via weak perspective projection, but it only acts as a weak constraint that is part of the overall objective function to be optimized. Therefore, its performance is highly influenced but not necessarily bounded by our derived error bounds.
Apart from purely weakly supervised methods, other end-to-end models with mixed supervision types have been established.
Kanazawa \etal~\cite{kanazawa18} try to estimate a parameterized 3D mesh of the human body in a generative fashion, while maintaining the image to 3D pose correspondence via direct 3D pose supervision and a weak perspective projection model. Similar to \cite{wandt19}, the projection model also acts as a weak constraint.
Finally, there are purely discriminative approaches with mixed 2D and 3D supervision. For example, Pavlakos \etal~\cite{pavlakos18} map an estimated volumetric 3D heatmap representation to the scaled and cropped 2D input image via weak perspective projection. And even without the weak perspective projection, the unaccounted input image normalization alone would introduce the same issues as a normalized perspective projection. Therefore, such architectures are again directly related to our study.

\textbf{Our work:} We focus on weakly and mixed supervised methods that utilize simplified projection models.
It is known that the objective function of monocular 3D human pose reconstruction is non-convex and difficult to optimize \cite{popa17}.
To the best of our knowledge, however, there is no existing study on lower error bounds in such approaches.

\section{Error Bounds of Projection Models}

Monocular 3D human pose estimation methods, either with 2D-to-3D pose uplifting or end-to-end image-based pipelines, often integrate a simplified projection model.
The main intention is to reduce the degrees of freedom and to avoid ambiguities in the 3D estimation process \cite{drover18, wandt19}.
We show that this induces a systematical error into the 3D estimates and derive theoretical lower bounds for this error.

For the evaluation of 3D HPE on benchmark datasets, the most commonly used evaluation metric is the \emph{Mean per-Joint Position Error} (MPJPE).
It is calculated as the mean \emph{Joint Position Error} (JPE; the Euclidean distances between ground truth and estimated joints) across all joints and poses.
With unknown object size, depth and possibly camera intrinsics, estimating a correct absolute position and scale of the 3D human pose is inherently difficult.
Common evaluation protocols therefore allow the alignment of 3D estimate and ground truth before calculating the MPJPE.
The most relaxed variant is the \emph{reconstruction error}, which allows the alignment of position, scale and rotation via Procrustes transformation.
In this work we assume a more strict evaluation protocol that only allows alignment in position and scale.
Specifically, we allow alignment of two designated root nodes and a per-subject scaling.
Our approach for scaling is as follows:
\begin{enumerate}[label={\arabic*.}]
	\item Calculate the average mean limb length $L_S$ of all poses of a  subject (person) $S$.
	\item For each subject $S$, scale the estimated 3D poses such that their mean limb lengths match $L_S$.
\end{enumerate}
This procedure is commonly known as \emph{Protocol 2} for evaluation on the Human3.6m dataset \cite{ionescu14}.
Identical or similar evaluation protocols are utilized in \cite{martinez17, zhou18, zhou16, tekin16, pavlakos17}.
We specifically derive lower bounds for the MPJPE w.r.t. Protocol 2.
Note that these minimal MPJPEs are optimistic best-case errors, such that the error bounds still hold for stricter protocols that only allow translation or no alignment at all.
The error bounds will simply be less tight.
Our derivations can be easily adjusted to better reflect stricter evaluation.

In the following sections, we work in the coordinate system of a standard perspective pin-hole camera with focal length $f$.
The camera is located at the origin of the coordinate system and is oriented towards positive $Z$ direction.
We regard the 2D-to-3D human pose estimator $G$ as a black box that, given a 2D input pose, predicts a reconstructed 3D pose such that its projection is identical to the input.
In practice, this requires the estimator to at least predict an absolute depth value for each joint of the pose \cite{drover18, pavlakos18}.
The resulting 3D estimate is then given by inverting the projection model.
In the following $P = [(X_1, Y_1, Z_1), \dots, (X_n, Y_n, Z_n)]$ denotes a 3D pose and $p = [(x_1, y_1), \dots, (x_n, y_n)]$ the corresponding 2D projection onto the image plane.
In general, we assume $f$ to be known.

\subsection{Normalized Perspective Projection}

In the normalized perspective projection model, the 2D pose is assumed to be normalized in position and scale. For 2D-to-3D pose uplifting, this necessitates translation and re-scaling of the 2D input pose.
We analyze the effects of both transformations on the final 3D estimate individually. \autoref{fig:normalized_perspective} gives an overview of our theoretical framework.

\subsubsection{Effects of Translation}
\label{sec:effects-of-translation}

First, we consider the normalization of input 2D poses by translating them to the origin of the coordinate system.
Note that this form of normalization is also part of many end-to-end 3D HPE pipelines that operate on cropped image patches.
Our basic setup for analyzing the effects of pose translation is as follows:
Let ground truth pose $P$ be centered at the origin of the X-Y-plane at depth $Z$.
Then the projection of joint $P_i$ is given by $p_i=(x_i, y_i)$, with
\begin{equation}
\label{eq:projected-point}
x_i = f \frac{X_i}{Z_i} \ ,\enspace y_i = f \frac{Y_i}{Z_i} \ .
\end{equation}
Now, we artificially shift $P$ by the vector $(dx, dy, 0)$ and project it onto the image plane.
After the resulting 2D pose is normalized by aligning it with the image plane's origin, it is re-projected into three dimensions by the 3D estimator $G$.
The resulting 3D pose $\widetilde{P}$ is then compared to $P$ via the MPJPE metric.
For simplicity we assume that $dy = 0$, that is, the pose $P$ is only shifted along the X-axis.

First, the artificial shift changes the X-component of the projected point $p_i$ to
\begin{equation}
\label{eq:projected-x-y}
x_i^\mathrm{S} = f \frac{X_i + dx}{Z_i} = x_i + f \frac{dx}{Z_i}\ .
\end{equation}
The projected pose is then normalized by shifting all projected points such that the root joint is located
in the origin of the image plane. 
Let the shifted root joint $P_r$ have coordinates $(dx, 0, Z)$.
Since $dy = 0$, normalization shifts each each point by $- f \frac{dx}{Z}$ along the X-axis.
The X-component  $\widetilde{x}_i$ of the normalized point is given by
\begin{equation}
\widetilde{x}_i
= x_i^\mathrm{S} - f \frac{dx}{Z}
= x_i + f dx \left (\frac{1}{Z_i} - \frac{1}{Z} \right) \ , 
\end{equation}
while the Y-component remains the same as in \autoref{eq:projected-point}.
After $G$ estimates the depth $\widetilde{Z}_i$ for each joint, the 2D points are re-projected into three dimensions using standard perspective projection.
The resulting coordinates are 
\begin{align}
\label{eq:re-projected-X}
\widetilde{X}_i &= \frac{\widetilde{x}_i}{f} \cdot \widetilde{Z}_i
= \widetilde{Z}_i \cdot \left( \frac{X_i}{Z_i} + dx \left( \frac{1}{Z_i} - \frac{1}{Z} \right) \right) \ , \\
\label{eq:re-projected-Y}
\widetilde{Y}_i &= \frac{y_i}{f} \cdot \widetilde{Z}_i = \frac{Y_i}{Z_i} \cdot \widetilde{Z}_i \ .
\end{align}
The JPE of the re-projected point is given by
\begin{equation}
\label{eq:delta-d}
\Delta d_i = \norm{ 
	\widetilde{P}_i - P_i
}_2 \ .
\end{equation}
With perfect estimation of $\widetilde{Z}_i = Z_i$, we have $\widetilde{Y}_i = Y_i$ and \autoref{eq:re-projected-X} describes a linear relationship
between the offset $\abs{dx}$ and the JPE $\Delta d_i$.
However, the JPE can actually become smaller for other $\widetilde{Z}_i \neq Z_i$. 
In order to obtain a correct lower bound for the JPE, we need to derive
the $\widetilde{Z}_i$ which minimizes $\Delta d_i$.
By substituting $a \coloneqq \left( \frac{X_i}{Z_i} + dx \left( \frac{1}{Z_i} - \frac{1}{Z} \right) \right )$
and $b \coloneqq \frac{Y_i}{Z_i}$, we have
\begin{align}
\label{eq:minimum-distance}
\Delta d_i^2 &= \left ( \widetilde{Z}_i a - X_i \right)^2 + \left ( \widetilde{Z}_i b - Y_i \right )^2 + \left ( \widetilde{Z}_i - Z_i \right )^2
\end{align}
This is a quadratic expression in $\widetilde{Z}_i$, whose global minimum is at
\begin{equation}
	\label{eq:optimal_z}
	\widetilde{Z}_i^\ast = \frac{a X_i + b Y_i + Z_i}{1 + a^2 + b^2}
  \end{equation}
Note that $\widetilde{Z}_i^\ast$ minimizes $\Delta d_i^2$ as well as the JPE $\Delta d_i$.
The detailed derivation can be found in the supplementary material.  
Given, $\widetilde{Z}_i^\ast$, the resulting $\widetilde{X}_i^\ast$ and $\widetilde{Y}_i^\ast$ follow from \autoref{eq:re-projected-X} and \autoref{eq:re-projected-Y}.
This  yields a total minimal error of 
	\begin{align}
	\Delta d_i &= 
	\sqrt{\left (\widetilde{X}_i^\ast - X_i \right )^2 + \left (\widetilde{Y}_i^\ast - Y_i \right )^2 + \left (\widetilde{Z}_i^\ast - Z_i \right )^2} \nonumber \\
	& = \abs{dx \left (\frac{1}{Z_i} - \frac{1}{Z} \right )} \sqrt{\frac{Y_i^2 + Z_i^2}{1 + a^2 + b^2}} \ . \label{eq:min_trans_error}
\end{align}
The minimal MPJPE for the whole pose can then be calculated as the mean of all $\Delta d_i$.
For a lower bound with $dx \neq 0$ \emph{and} $dy \neq 0$, $b$ has to be substituted with $\left( \frac{Y_i}{Z_i} + dy \left( \frac{1}{Z_i} - \frac{1}{Z} \right) \right )$ in the above equations.

The result shows that even with the system estimating the depth of each point optimally,
the estimated 3D pose will not match the original 3D pose.
The effects of 2D pose normalization by translation only vanish if $dx = 0$ or $Z_i = Z$ for all $i$. 
In the latter case all joints of the pose are located in a plane parallel to the image plane.


During evaluation with Protocol 2, the estimated 3D pose is aligned to the ground truth by shifting and scaling.
For the root joint $r$ we have $Z_r = Z$ in \autoref{eq:min_trans_error}, so $\Delta d_r = 0$. 
This means $(\widetilde{X}_r, \widetilde{Y}_r, \widetilde{Z}_r) = (X_r, Y_r, Z_r)$, that is,
the root joints of both poses are already aligned.
Additionally, Protocol 2 calculates a per-pose scaling factor $\alpha_{\widetilde{P}}$ such that the average limb length in $\alpha_{\widetilde{P}} \cdot \widetilde{P}$ is equal to the average limb length of the respective subject.
Thus, for experimental evaluation, the pose $\widetilde{P}$ has to be re-scaled
by $\alpha_{\widetilde{P}}$ in all dimensions before the MPJPE is calculated.

\subsubsection{Effects of Scaling}
\label{sec:z-shift-error}
It is common for many 3D HPE architectures to normalize the 2D input to match a fixed, but otherwise arbitrary scale. 
For 2D-to-3D uplifting, the 2D input pose is scaled to match \eg a specific standard deviation or torso size \cite{drover18, wandt19}. 
In image-based pipelines, this corresponds to resizing the input image to a specific scale.
For a fixed 3D human pose, the scale of the 2D projection depends on the focal length $f$ and the absolute depth $Z$.
Since we assume $f$ to be known, we analyze the effect of $Z$ on the best-case 3D estimate.



Again consider points $P_i=(X_i, Y_i, Z_i) \in P$.
Assume that $P$ is shifted by $dz$ along the Z-axis.
The coordinates of the projected points on the image plane are
\begin{align}
x_i = f \frac{X_i}{Z_i + dz} \ ,\enspace
y_i = f \frac{Y_i}{Z_i + dz} \ .
\end{align}
Due to the Z-shift, the projected points now need to be scaled by $\rho$ such that they match the fixed target scale.
After $G$ estimates the depths $\widetilde{Z}_i$, each point is re-projected into three dimensional space:
\begin{align}
\widetilde{X}_i &= \frac{\rho \cdot x_i}{f} \cdot \widetilde{Z}_i
= \rho \cdot \frac{X_i}{Z_i + dz} \cdot \widetilde{Z}_i \ , \label{eq:x_i-min}\\
\widetilde{Y}_i &= \frac{\rho \cdot y_i}{f} \cdot \widetilde{Z}_i
= \rho \cdot \frac{Y_i}{Z_i + dz}\cdot \widetilde{Z}_i \ . \label{eq:y_i-min}
\end{align}
To obtain a best-case 3D estimate, we minimize the JPE with respect to $\widetilde{Z}_i$.
Following \autoref{eq:delta-d}, the squared JPE for joint $P_i$ is given by
\begin{align}
\Delta d_i^2 &= \left ( \widetilde{Z}_i a - X_i \right)^2 + \left ( \widetilde{Z}_i b - Y_i \right)^2 + \left ( \widetilde{Z}_i - Z_i \right)^2,
\end{align}
with $a \coloneqq \rho \cdot \frac{X_i}{Z_i + dz}$ and $b \coloneqq \rho \cdot \frac{Y_i}{Z_i + dz}$.
Minimizing $\Delta d_i^2$ w.r.t. $\widetilde{Z}_i$ yields the same expression as in \autoref{eq:minimum-distance}, with 
\begin{equation}
\widetilde{Z}_i^\ast = \frac{a X_i + b Y_i + Z_i}{1 + a^2 + b^2} \ .
\end{equation}
The detailed derivation is again part of the supplementary material.
With $\widetilde{X}_i^\ast$ and $\widetilde{Y}_i^\ast$ from Equations~\eqref{eq:x_i-min} and \eqref{eq:y_i-min}, the minimal JPE is given by 
\begin{equation}
\label{eq:z-shift-error}
\Delta d_i = \abs{1 - \rho \cdot \frac{Z_i}{Z_i + dz}}  \sqrt{\frac{X_i^2 + Y_i^2}{1 + a^2 + b^2}}\ .
\end{equation}
The MPJPE $\Delta d$ of the whole pose $P$ can be calculated as the mean of all $\Delta d_i$.
Similar to \autoref{sec:effects-of-translation}, during experimental evaluation, scaling and shifting according to Protocol 2 is applied.

The reason for the existence of an additional MPJPE is the scale factor $\rho$.
As $\rho$ is calculated for the whole pose rather than for each joint separately,
$\rho \cdot \frac{Z_i}{Z_i + dz}$ is usually \emph{not} equal to 1.
This is only the case if $\rho = \frac{Z_i + dz}{Z_i}$ for all $i$, which means $Z_i = Z_j$ for all $1 \leq i, j \leq n$.
In this case, all joints are located on the same X-Y-plane and the additional MPJPE vanishes.


\subsection{Weak Perspective Projection}

\begin{figure}[t]
  \centering
  \includegraphics[width=0.8\linewidth]
                   {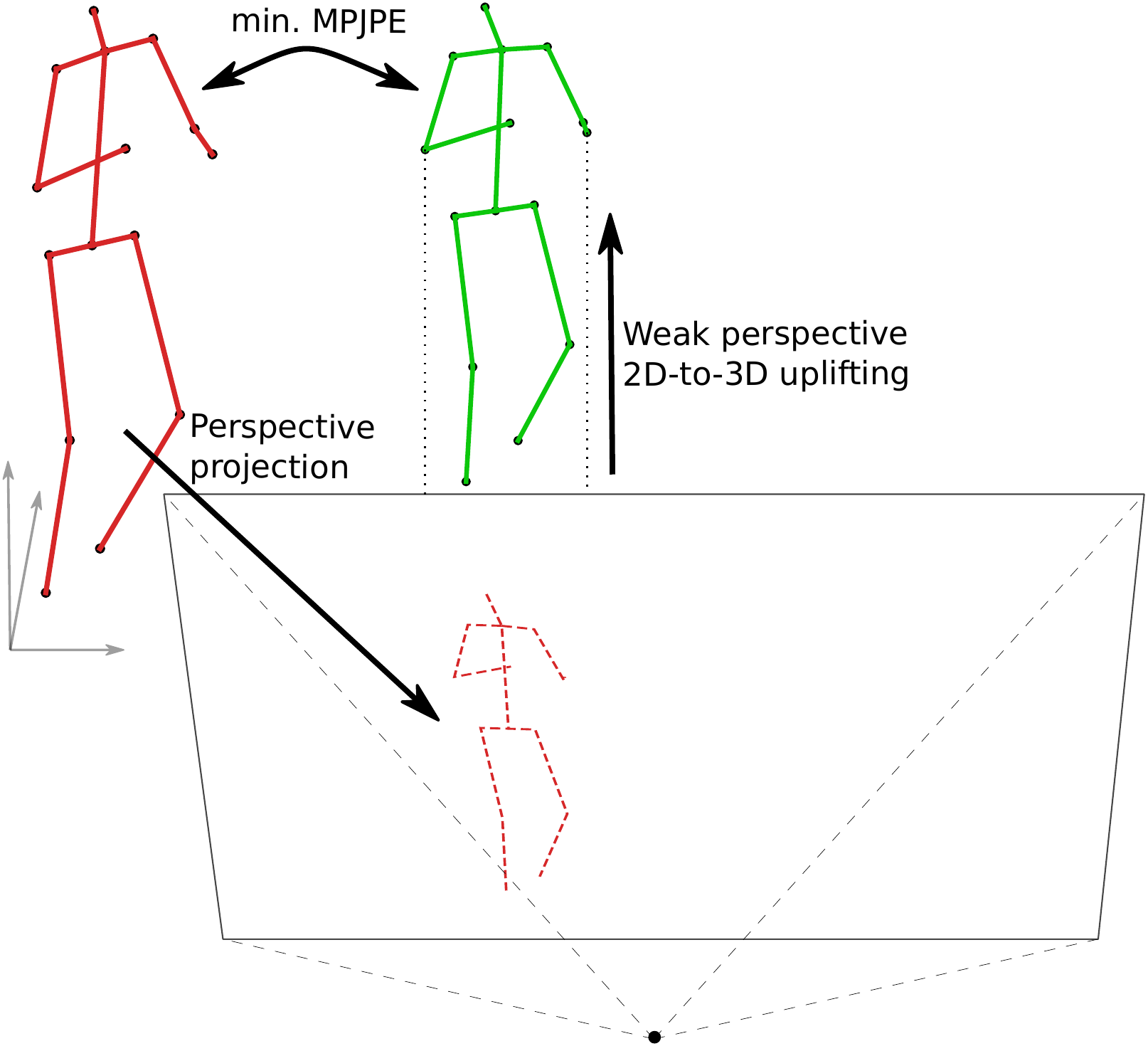}
  \caption{Lower bound MPJPE for weak perspective projection: We derive the best-case 2D-to-3D re-projection under this projection model and compare it to the 3D ground truth.
  }
  \label{fig:weak_perspective}
\end{figure}

Instead of retaining a true perspective projection with normalized 2D inputs, many weakly and mixed supervised methods adopt the simpler weak perspective projection model.
We can again derive an optimal 2D-to-3D human pose estimate under this projection model and, by comparing it to the 3D ground truth, establish a lower error bound. \autoref{fig:weak_perspective} visualizes the conceptual setup.

Assume we have an arbitrary ground truth 3D pose $P$ with joints $P_i = (X_i, Y_i, Z_i)$ and its 2D perspective projection $p$ with joints $p_i = (x_i, y_i)$.
The weak perspective projection model is structurally invariant to shifts perpendicular to the image plane.
We therefore simplify our derivations by shifting $p$ into the origin, resulting in $p^\prime$.
With $p^\prime$ as its input, the 3D estimator $G$ can now predict the points in 3D space given as $\widetilde{P}_i = (\widetilde{X}_i, \widetilde{Y}_i, \widetilde{Z}_i)$.
In order to reflect the X-Y shift in the input $p^\prime$, the goal is to reconstruct a equally X-Y centered version of $P$, denoted as $P^\prime$.
If the estimate matches $P^\prime$, it will equally match the original 3D pose $P$ during Protocol 2 alignment.
The 3D estimate has to adhere to a weak perspective projection, with
\begin{align}
	x^\prime_i = f \frac{\widetilde{X}_i}{Z_{\text{avg}}} \ ,\enspace
	y^\prime_i = f \frac{\widetilde{Y}_i}{Z_{\text{avg}}} .
\end{align}
Even though $Z_{\text{avg}}$ is conceptually the average depth over all keypoints, it can be subsumed with $f$ into an independent scaling factor $s = \frac{Z_{\text{avg}}}{f}$.
The 3D estimator has to predict $s$ alongside the individual depths $\widetilde{Z}_i$ for all keypoints \cite{kanazawa18, wandt19}.
With $\widetilde{Z}_i$ independent from $s$, the weak perspective model allows a best case depth estimate with $\widetilde{Z}_i = Z^\prime_i$.
We can now derive the optimal scaling $s$ that minimizes the overall difference between $\widetilde{P}$ and $P^\prime$. For a closed-form solution, we minimize the mean squared per-joint error $\Delta d_2$ instead of the non-squared MPJPE $\Delta d$:
\begin{align}
	\Delta d_2(s) &= \frac{1}{n}\sum_i \norm{ 
	\begin{pmatrix}
	 X^\prime_i - sx^\prime_i \\
	 Y^\prime_i - sy^\prime_i\\
	Z^\prime_i - \widetilde{Z}_i
	\end{pmatrix}
}_2^2 \ .
\end{align}
Under optimal depth estimation, $\Delta d_2$ is reduced to
\begin{align}
	\Delta d_2(s) &= \frac{1}{n} \sum_i \left( X^\prime_i - sx^\prime_i \right )^2 + \left( Y^\prime_i - sy^\prime_i \right )^2.
\end{align}
This quadratic expression is minimal at
\begin{align}
	s^* &= \arg \min_s \Delta d_2(s) = \frac{\sum_i X^\prime_ix^\prime_i + Y^\prime_iy^\prime_i}{\sum_i x_i^{\prime 2} + y_i^{\prime 2}}.
\end{align}
The detailed derivation can again be found in the supplementary material.
Note that in general, $s^* \neq \arg \min_s \Delta d(s)$, \ie the obtained scale does not necessarily minimize the non-squared MPJPE.
However, we have empirically found that using $s^*$ only leads to an increase of $\Delta d(s)$ by less then $0.2$mm compared to the actual optimal scaling on our evaluation datasets.

\subsection{Relaxed Normalized Perspective Projection}
\label{sec:modified-network}
In order to experimentally validate the derived lower bounds and to show how to circumvent them, we evaluate and modify a baseline 3D HPE architecture.
For this we consider the baseline 2D-to-3D HPE system by Drover \etal \cite{drover18}.
It is based on a weakly supervised GAN architecture \cite{goodfellow14} in which the generator receives 2D poses normalized in position and scale and estimates the depth of each joint.
The 2D pose is then re-projected into 3D following standard perspective projection.

We re-implement this architecture with minor modifications:
We remove batch normalization, as it proved detrimental for training stability and convergence and apply adaptive gradient clipping \cite{chorowski14}.
Apart from that, we use the same hyper-parameters.

Our experiments show that the effects of scaling might be negligible in most applications.
Meanwhile, normalization by translation introduces a major increase in MPJPE.
Thus, we focus on how to avoid the adverse effects of translation.
In \autoref{eq:re-projected-X}, if the 2D points would be re-projected correctly (\ie shifted back before re-projection), the lower bound in \autoref{eq:delta-d} would no longer hold true.
For this, assume the generator estimated all depths $\widetilde{Z}_i$ perfectly, that is, $\widetilde{Z}_i = Z_i$.
If the 2D pose is shifted back and then re-projected into 3D, we have $\widetilde{X}_i = X_i$ and $\widetilde{Y}_i = Y_i$.
In this case, the re-projected pose is now equal to the ground truth pose.

We adapt the baseline system by replacing the full normalization of the input 2D pose with two relaxed variants:
In the first variant, the 2D input is still normalized in position and scale, but the re-projection into 3D space is performed on the non-shifted 2D input (thus invalidating the lower bound).
For the second variant, the normalization by translation is removed entirely.
This allows the 3D estimator to condition its per-joint depth estimate on the actual, non-shifted 2D joint locations.
In both cases, the system is explicitly trained on uniformly shifted 2D input poses.

\section{Experimental Results}

We first show how the derived lower bounds for the normalized perspective projection change with varying offsets of artificially shifted poses.
We validate them by comparing them to the baseline 2D-to-3D pose estimation system  in \cite{drover18}.
Additionally, we evaluate the proposed relaxed normalization when integrated into the same 3D pose estimator.
Finally, in order to quantify the theoretical error bounds' impact, we evaluate them for the normalized and weak perspective projection on the test sets of three commonly used 3D HPE benchmark datasets, namely Human3.6m \cite{ionescu14}, MPI-INF-3DHP \cite{mehta17} and CMU Panoptic \cite{joo15}.
Note that in all experiments, we assume perfect 2D poses as the input to 2D-to-3D uplifting.
All results are evaluated according to Protocol 2, unless stated otherwise.

\emph{Human3.6m} \cite{ionescu14} consists of 3D human poses captured by a MoCap system and four RGB cameras in an indoor lab setting.
We use the common split with subjects S9 and S11 for evaluation and the remaining subjects for training and employ the same 14 joint model as \cite{drover18} with the addition of the pelvis.
Most of our experiments are conducted on this dataset, as it is the largest and most commonly used benchmark for single-person 3D HPE.
Similarly, \emph{MPI-INF-3DHP} \cite{mehta17} contains 3D human poses from different activities, but with more viewpoint and pose variety.
We use the pre-defined split and evaluate on all test subjects.
Finally, \emph{CMU Panoptic} \cite{joo15} is the currently largest multi-person 3D HPE dataset.
The poses are captured by 30 HD cameras arranged in a sphere-like structure.
We treat the poses of all subjects individually and adopt the evaluation protocol from \cite{zanfir18, fabbri2020}, excluding the no longer maintained "Mafia" sequence.

\subsection{Translation}
\label{sec:translation-experimental}

\begin{figure}[t]
		\centering
		\begin{minipage}{\columnwidth}
			\centering
			\begin{tikzpicture}[]
				\begin{axis}[
					small,
					width=\textwidth,
					xlabel={$dx$ [m]},
					y label style={at={(axis description cs:0.24,1.08)},rotate=-90,anchor=north},
					ylabel={MPJPE [mm]},
					grid=both,
					enlarge x limits = false,
					grid style={draw=gray!70, line width=.5pt},
					minor tick num = 1,
					tick style={draw=gray!50},
					xtick={-7, ..., 7},
					ymin=-5,
					extra x ticks={0},
					scaled x ticks = false,
					no markers,
					every axis plot/.append style={}
					]
					\addplot +[restrict expr to domain={\coordindex}{120:461}] table [x=a, y=b, col sep=comma] {figures/plot_e_03_07_original_x_shift.csv};
					\addplot +[restrict expr to domain={\coordindex}{120:461}] table [x=a, y=b, col sep=comma] {figures/plot_theoretical_x_shift_scaled.csv};
				\end{axis}
			\end{tikzpicture}
		\end{minipage}
		\caption{Theoretical lower bound (red) and experimental (blue) MPJPE with \cite{drover18} on Human3.6m poses at different offsets $dx$.
		The theoretical error evaluates the best-case 3D estimates under normalized perspective projection.}
	\label{fig:x-shift-error}
\end{figure}
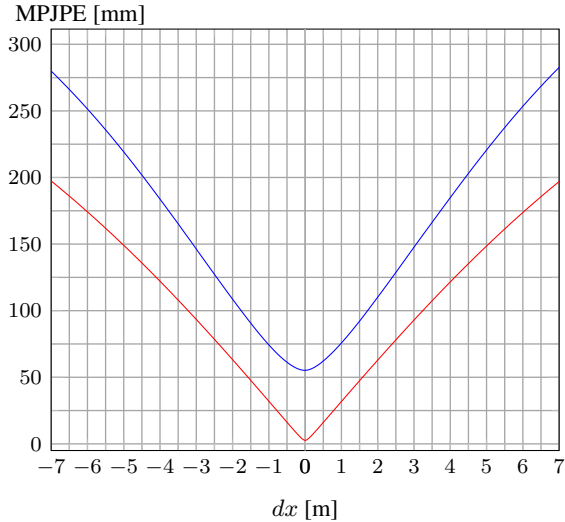

First, we analyze the effect of translation on the normalized perspective lower bound error and compare it to the observed performance of the corresponding baseline system in \cite{drover18}, trained on Human3.6m.
We use all 3D test set poses in the Human3.6m dataset, center them in the X-Y plane and place them on the Z axis such that the resulting 2D projections already satisfy the target scale of the baseline system.
This results in an average camera distance of 5m to the poses. We then artificially shift the 3D poses along the X axis by offset $dx$.
After projecting them onto the image plane, we calculate the best-case 3D estimates following \autoref{eq:optimal_z} and evaluate them according to Protocol 2.
Additionally, we feed the synthetic 2D poses into the baseline 3D pose estimation system.
The results with respect to $dx$ are depicted in \autoref{fig:x-shift-error}.

Starting with its minimum at $dx=0$m, the theoretical lower bound error quickly increases nearly linearly in the depicted interval, with a minimal MPJPE of already $62.9$mm at only $dx=2$m offset.
At an average depth of $5$m, this reflects a still very common camera viewpoint, where the human pose is observed at an angle of $22^\circ$.
The results of the baseline 2D-to-3D pose estimator show similar characteristics.
Starting at an MPJPE of $55$mm at $dx=0$m, it increases similarly with a doubled MPJPE at only $2$m offset. For larger offsets, the error increases even faster than the lower bound error.
Overall, the results show that normalization by translation leads to a quickly increasing lower bound error, which can be equally observed for the baseline 3D estimator.

\subsection{Scaling}
\label{sec:error-on-shift-experimental}

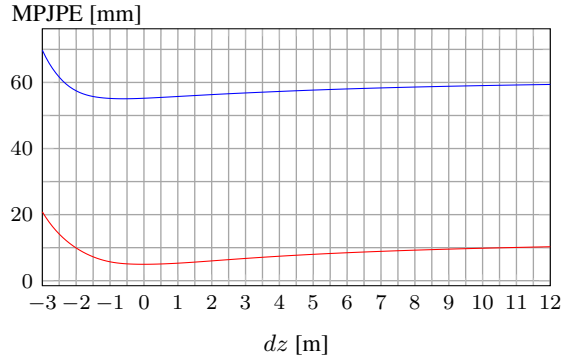
\begin{figure}[t]
		\centering
		\begin{tikzpicture}[]
		\begin{axis}[
		small,
		width=\columnwidth,
		height=5cm,
		xlabel={$dz$ [m]},
		y label style={at={(axis description cs:0.25,1.13)},rotate=-90,anchor=north},
		ylabel={MPJPE [mm]},
		grid=both,
		enlarge x limits = false,
		grid style={draw=gray!70,, line width=.5pt},
		minor tick num = 1,
		tick style={draw=gray!50},
		scaled x ticks = false,
		no markers,
		xtick={-3,-2,...,15},
		every axis plot/.append style={}
		]
		\addplot +[restrict expr to domain={\coordindex}{0:150}] table [x=a, y=b, col sep=comma] {figures/plot_e_03_07_original_z_shift.csv};
		\addplot[color=red, restrict expr to domain={\coordindex}{0:150}] table [x=a, y=b, col sep=comma] {figures/plot_theoretical_z_shift_scaled.csv};
		\end{axis}
		\end{tikzpicture}
	\caption{Theoretical lower bound (red) and experimental (blue) MPJPE with \cite{drover18} on Human3.6m poses at different depth offsets $dz$.
		The depth at $dz=0$ is approximately 5m.
	}
	\label{fig:z-shift-error}
\end{figure}

We use the same experimental setup to evaluate the theoretical lower bound error that is introduced by scale normalization.
This time, however, we artificially shift the 3D poses by varying offsets $dz$ along the optical axis.
This results in 2D projections that need to be normalized in scale to match the target scale of the baseline system.
We evaluate the theoretical error as well as the change in MPJPE when applying the baseline system. The results are depicted in \autoref{fig:z-shift-error}.

The behavior of the theoretical error and the baseline system again show similar characteristics.
Compared to the effect of shifting along the X axis, the MPJPE does not grow significantly for changes in $dz$.
Only when poses are moving closer to the camera we observe a notable MPJPE increase, with $20$mm at $-3$m offset.
This results from the fact that with smaller camera distances, projection ray angles change more severely.
The reason for the $5$mm theoretical MPJPE at $dz = 0$ stems from Protocol 2 evaluation.
As there is intra-subject variance in average limb lengths in Human3.6m, this minor error is introduced when scaling the estimated poses.

In general, the results show that normalization by scaling does introduce a systematic error. 
Apart from the case of poses being very close to the camera, the increase in MPJPE might be negligible in most applications.

\subsection{Relaxed Normalization}

Given the notable influence of normalization by translation, we evaluate our proposed relaxed normalization with the same experimental setup as in \autoref{sec:translation-experimental}. Again, the baseline as well as the proposed variants are trained on Human3.6m.

\begin{figure}[t]
		\centering
		\begin{tikzpicture}[trim axis left]
		\begin{axis}[
		small,
		width=\columnwidth,
		height=\axisdefaultheight,
		xlabel={$dx$ [m]},
		y label style={at={(axis description cs:0.24,1.08)},rotate=-90,anchor=north},
		ylabel={MPJPE [mm]},
		grid=both,
		enlarge x limits = false,
		grid style={draw=gray!70,line width=.5pt},
		minor tick num = 1,
		tick style={draw=gray!50},
		xtick={-7,...,7},
		ymin=-5,
		extra x ticks={0},
		scaled x ticks = false,
		no markers,
		every axis plot/.append style={}
		]
		\addplot +[restrict expr to domain={\coordindex}{120:461}] table [x=a, y=b, col sep=comma] {figures/plot_e_03_07_original_x_shift.csv};
		\addplot[color=red, restrict expr to domain={\coordindex}{30:371}] table [x=a, y=b, col sep=comma] {figures/plot_e_26_08_shifted_generator_x_shift.csv};
		\addplot[color=black!60!green, restrict expr to domain={\coordindex}{30:371}] table [x=a, y=b, col sep=comma] {figures/plot_e_11_08_default_generator_x_shift.csv};
		\end{axis}
		\end{tikzpicture}
	\caption{
      Comparison of the baseline \cite{drover18} (blue) and our modified version with relaxed normalization for Human3.6m poses at different offsets $dx$: correct re-projection (green); without normalization by translation (red).
	}
	\label{fig:x-shift-error-correct-reproj}
\end{figure}
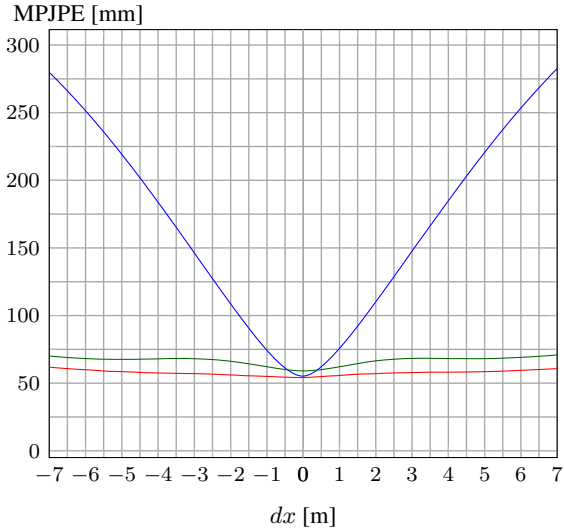

The results for varying offsets $dx$ are depicted in \autoref{fig:x-shift-error-correct-reproj}. With the re-projection into 3D space based on the original 2D input, the MPJPE increases much slower with respect to $dx$ and clearly surpasses the lower error bound in \autoref{fig:x-shift-error}.
Compared to the baseline, we observe a negative effect only for very small offsets $dx$.
Avoiding normalization by translation entirely leads to another reduction of the MPJPE by $10$mm across all offsets.
At $dx=0$ the performance now equals the baseline system. 
This confirms the conjecture that the depth estimates need to be conditioned on the offsets $dx$ to avoid ambiguities in the 3D estimates.
At a large offset of $dx=7$m, the relaxed normalization leads to an increase in MPJPE of only $6.6$mm, compared to the $227.4$mm for the baseline.
Despite no normalization by translation, we do not observe any divergence issues.

We also evaluate on the actual Human3.6m test set, \ie the 3D and 2D poses as observed by the camera setup.
\autoref{tbl:lower-bounds} (top) shows the resulting MPJPE under Protocol 2 for the baseline system and our proposed modification with relaxed normalization.
At $50.9$mm, our approach improves upon the baseline by notable $15.4$mm.
Despite the limited variability in the Human3.6m poses and viewpoints, the baseline is still heavily influenced by translation normalization.
We additionally evaluate the less strict reconstruction error, where we observe identical results for the baseline as well as our relaxed normalization variant.
The systematic error of translation normalization for 2D-to-3D uplifting is compensated by the additional orientation alignment during evaluation.
For applications where pose orientation relative to the camera is irrelevant, the effects of translation are thus far less severe.

Overall, our simple modifications to an otherwise unchanged 2D-to-3D uplifting model can avoid the adverse effects of normalization almost entirely.
Our proposal can be integrated into every weakly supervised system following the same normalized perspective projection.

\subsection{Lower Error Bounds on 3D HPE Datasets}

\begin{table}[t]
  \centering
  
  \begin{tabularx}{0.8\columnwidth}{ l *{2}{Y}}
    \toprule
    & Protocol 2 & Reconstruction error\\
    \midrule
    \cite{drover18} & 66.3 & 39.0\\
    Ours & 50.9 & 39.0\\
    \bottomrule
  \end{tabularx}
  
  \vspace{0.5em}
  
  \begin{tabularx}{\columnwidth}{ l *{3}{Y}}
    \toprule
    &Human 3.6m &MPI-INF-3DHP &Panoptic\\
    \midrule
    State-of-the-art & 49.6 \cite{sun18}$^*$ & 101.5 \cite{rhodin18} & 60.7 \cite{fabbri2020}$^*$ \\
    \midrule
    Norm. persp. & 19.3 & 52.0 & 54.7  \\
    Weak persp. &  20.6 & 47.8 & 47.2 \\
    \bottomrule
  \end{tabularx}

  \caption{Top: Results of 2D-to-3D uplifting with relaxed normalization on Human3.6m. Bottom:
    Comparison of state-of-the-art results and our theoretical lower bounds evaluated with Protocol 2 on common 3D HPE benchmark datasets. All results are given in MPJPE (mm). $^*$ use stricter evaluation without scale alignment.
  }
  \label{tbl:lower-bounds}
\end{table}

\begin{figure}[t]
  \centering
  \includegraphics[width=0.98\linewidth]
                   {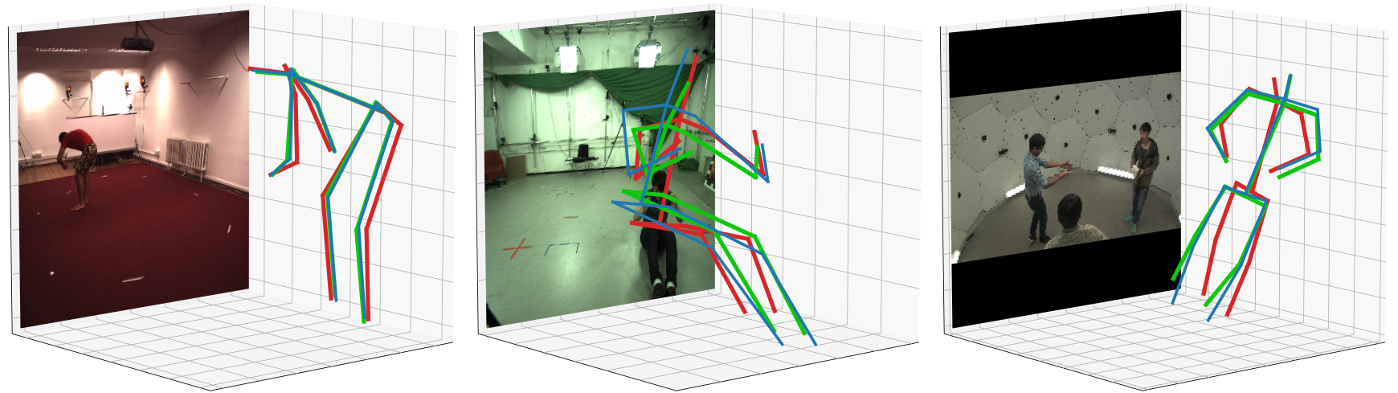}
  \caption{Exemplary ground truth poses (red) from Human3.6m, MPI-INF-3DHP and CMU Panoptic (left to right). The best-case 3D estimates under normalized and weak perspective projection are depicted in blue and green. 
  }
  \label{fig:qualitative_examples}
\end{figure}

Finally, \autoref{tbl:lower-bounds} (bottom) shows the evaluated lower bound errors for the normalized and the weak perspective projection model on the test set poses of the most common 3D HPE benchmarks.
For the normalized perspective variant, we only assume normalization by translation due to its considerable negative effect. 
Additionally, \autoref{fig:qualitative_examples} depicts qualitative examples of the best-case pose estimates under those projection models. We provide additional examples in the supplementary material.

Across all datasets, both projection models lead to a similar lower bound MPJPE.
For Human3.6m, the minimal error is relatively small, at around $20$mm.
For MPI-INF-3DHP and CMU Panoptic, however, the minimal error is at approximately $50$mm already.
Compared to Human3.6m, those datasets show a much wider variety in camera viewpoints, with poses being off-center, closer to the camera and with more absolute depth variation.
This seems to affect the best-case 3D estimates with normalized and weak perspective projection to the same degree.
Since the latter datasets better reflect in-the-wild human 3D poses, the effect of simplified projection models on 3D HPE applications is considerable.
For reference, we report current state-of-the-art results on the respective datasets.
Note that these results are obtained with fully supervised discriminative methods that do not employ simplified projection models.
They are not limited by our derived lower bounds.
Still, when comparing the results, the derived minimal errors are already at $50\%$ or more of the state-of-the-art MPJPE results.
Note that the minimal errors assume otherwise maximally optimistic 3D estimates with perfect 2D input poses.
Therefore, any approach that employs 2D-to-3D uplifting with either normalized or weak perspective projection will hardly be able to surpass the current state-of-the-art.

\section{Conclusion}

We presented a theoretical analysis of simplified projection models that are used in current weakly supervised 2D-to-3D HPE uplifting methods.
We derived lower bound errors under a common evaluation protocol and metric by providing closed-form solutions for the best-case 3D pose estimates.
Our analysis showed that 2D pose normalization by translation as well as the weak perspective projection model introduce a considerable error, even under overly-optimistic conditions.
We showed how a relaxed normalization can be directly integrated into a baseline architecture to overcome the otherwise guaranteed minimal error.
The evaluation on real-world 3D HPE datasets provides a baseline for future comparison between weakly and fully supervised methods.
The comparison to state-of-the-art results showed that it is crucial for any weakly supervised architecture to avoid simplified projection models in order to achieve competitive results on current benchmarks.
Our proposed relaxed normalization is one possibility to consider.

\pagebreak
\bibliographystyle{ieee}
\bibliography{bibliography}

\begin{thebibliography}{10}\itemsep=-1pt

\bibitem{chen17}
C.-H. Chen and D.~Ramanan.
\newblock {3D Human Pose Estimation = 2D Pose Estimation + Matching}.
\newblock In {\em Proceedings of the IEEE Conference on Computer Vision and
  Pattern Recognition (CVPR)}, July 2017.

\bibitem{chorowski14}
J.~Chorowski, D.~Bahdanau, K.~Cho, and Y.~Bengio.
\newblock {End-to-end Continuous Speech Recognition using Attention-based
  Recurrent NN: First Results}.
\newblock In {\em {NIPS 2014 Workshop on Deep Learning}}, Dec. 2014.

\bibitem{drover18}
D.~Drover, R.~MV, C.-H. Chen, A.~Agrawal, A.~Tyagi, and C.~Phuoc~Huynh.
\newblock {Can 3D Pose be Learned from 2D Projections Alone?}
\newblock In {\em Proceedings of the European Conference on Computer Vision
  (ECCV) Workshops}, September 2018.

\bibitem{fabbri2020}
M.~Fabbri, F.~Lanzi, S.~Calderara, S.~Alletto, and R.~Cucchiara.
\newblock {Compressed Volumetric Heatmaps for Multi-Person 3D Pose Estimation}.
\newblock In {\em Proceedings of the IEEE/CVF Conference on Computer Vision and
  Pattern Recognition}, pages 7204--7213, 2020.

\bibitem{tung17}
H.-Y. Fish~Tung, A.~W. Harley, W.~Seto, and K.~Fragkiadaki.
\newblock {Adversarial Inverse Graphics Networks: Learning 2D-To-3D Lifting and
  Image-To-Image Translation From Unpaired Supervision}.
\newblock In {\em Proceedings of the IEEE International Conference on Computer
  Vision (ICCV)}, Oct 2017.

\bibitem{goodfellow14}
I.~J. Goodfellow, J.~Pouget-Abadie, M.~Mirza, B.~Xu, D.~Warde-Farley, S.~Ozair,
  A.~Courville, and Y.~Bengio.
\newblock {Generative Adversarial Networks}.
\newblock In {\em {Advances in Neural Information Processing Systems 27}},
  pages 2672--2680. June 2014.

\bibitem{grinciunaite16}
A.~Grinciunaite, A.~Gudi, E.~Tasli, and M.~den Uyl.
\newblock {Human Pose Estimation in Space and Time using 3D CNN}.
\newblock In {\em {Computer Vision -- ECCV 2016 Workshops}}, Oct. 2016.

\bibitem{ionescu14}
C.~Ionescu, D.~Papava, V.~Olaru, and C.~Sminchisescu.
\newblock {Human3.6M: Large Scale Datasets and Predictive Methods for 3D Human
  Sensing in Natural Environments}.
\newblock {\em IEEE Transactions on Pattern Analysis and Machine Intelligence},
  36(7):1325--1339, Jul 2014.

\bibitem{joo15}
H.~Joo, H.~Liu, L.~Tan, L.~Gui, B.~Nabbe, I.~Matthews, T.~Kanade, S.~Nobuhara,
  and Y.~Sheikh.
\newblock {Panoptic Studio: A Massively Multiview System for Social Motion
  Capture}.
\newblock In {\em {The IEEE International Conference on Computer Vision
  (ICCV)}}, 2015.

\bibitem{kanazawa18}
A.~Kanazawa, M.~J. Black, D.~W. Jacobs, and J.~Malik.
\newblock {End-to-End Recovery of Human Shape and Pose}.
\newblock In {\em Proceedings of the IEEE Conference on Computer Vision and
  Pattern Recognition (CVPR)}, June 2018.

\bibitem{kanazawa19}
A.~Kanazawa, J.~Y. Zhang, P.~Felsen, and J.~Malik.
\newblock {Learning 3d human dynamics from video}.
\newblock In {\em Proceedings of the IEEE Conference on Computer Vision and
  Pattern Recognition}, pages 5614--5623, 2019.

\bibitem{kocabas2020}
M.~Kocabas, N.~Athanasiou, and M.~J. Black.
\newblock {VIBE: Video inference for human body pose and shape estimation}.
\newblock In {\em Proceedings of the IEEE/CVF Conference on Computer Vision and
  Pattern Recognition}, pages 5253--5263, 2020.

\bibitem{kocabas2019}
M.~Kocabas, S.~Karagoz, and E.~Akbas.
\newblock {Self-supervised learning of 3d human pose using multi-view
  geometry}.
\newblock In {\em Proceedings of the IEEE Conference on Computer Vision and
  Pattern Recognition}, pages 1077--1086, 2019.

\bibitem{martinez17}
J.~Martinez, R.~Hossain, J.~Romero, and J.~J. Little.
\newblock {A simple yet effective baseline for 3d human pose estimation}.
\newblock {\em 2017 IEEE International Conference on Computer Vision (ICCV)},
  Aug. 2017.

\bibitem{mehta17}
D.~Mehta, H.~Rhodin, D.~Casas, P.~Fua, O.~Sotnychenko, W.~Xu, and C.~Theobalt.
\newblock {Monocular 3D Human Pose Estimation In The Wild Using Improved CNN
  Supervision}.
\newblock {\em 2017 International Conference on 3D Vision (3DV)}, Oct. 2017.

\bibitem{nibali2019}
A.~Nibali, Z.~He, S.~Morgan, and L.~Prendergast.
\newblock {3d human pose estimation with 2d marginal heatmaps}.
\newblock In {\em 2019 IEEE Winter Conference on Applications of Computer
  Vision (WACV)}, pages 1477--1485. IEEE, 2019.

\bibitem{pavlakos18}
G.~Pavlakos, X.~Zhou, and K.~Daniilidis.
\newblock {Ordinal Depth Supervision for 3D Human Pose Estimation}.
\newblock In {\em Proceedings of the IEEE Conference on Computer Vision and
  Pattern Recognition (CVPR)}, June 2018.

\bibitem{pavlakos17}
G.~Pavlakos, X.~Zhou, K.~G. Derpanis, and K.~Daniilidis.
\newblock {Coarse-to-Fine Volumetric Prediction for Single-Image 3D Human
  Pose}.
\newblock {\em 2017 IEEE Conference on Computer Vision and Pattern Recognition
  (CVPR)}, pages 1263--1272, 2016.

\bibitem{pavllo19}
D.~Pavllo, C.~Feichtenhofer, D.~Grangier, and M.~Auli.
\newblock {3D Human Pose Estimation in Video With Temporal Convolutions and
  Semi-Supervised Training}.
\newblock In {\em The IEEE Conference on Computer Vision and Pattern
  Recognition (CVPR)}, June 2019.

\bibitem{popa17}
A.-I. Popa, M.~Zanfir, and C.~Sminchisescu.
\newblock {Deep Multitask Architecture for Integrated 2D and 3D Human Sensing}.
\newblock In {\em Proceedings of the IEEE Conference on Computer Vision and
  Pattern Recognition (CVPR)}, July 2017.

\bibitem{rhodin18}
H.~Rhodin, J.~Sp{\"o}rri, I.~Katircioglu, V.~Constantin, F.~Meyer,
  E.~M{\"u}ller, M.~Salzmann, and P.~Fua.
\newblock {Learning Monocular 3D Human Pose Estimation From Multi-View Images}.
\newblock In {\em Proceedings of the IEEE Conference on Computer Vision and
  Pattern Recognition (CVPR)}, June 2018.

\bibitem{rogez16}
G.~Rogez and C.~Schmid.
\newblock {MoCap-guided Data Augmentation for 3D Pose Estimation in the Wild}.
\newblock In D.~D. Lee, M.~Sugiyama, U.~V. Luxburg, I.~Guyon, and R.~Garnett,
  editors, {\em Advances in Neural Information Processing Systems 29}, pages
  3108--3116. Curran Associates, Inc., 2016.

\bibitem{sun18}
X.~Sun, B.~Xiao, F.~Wei, S.~Liang, and Y.~Wei.
\newblock {Integral Human Pose Regression}.
\newblock In {\em Proceedings of the European Conference on Computer Vision
  (ECCV)}, September 2018.

\bibitem{tekin16}
B.~Tekin, A.~Rozantsev, V.~Lepetit, and P.~Fua.
\newblock {Direct Prediction of 3D Body Poses from Motion Compensated
  Sequences}.
\newblock {\em 2016 IEEE Conference on Computer Vision and Pattern Recognition
  (CVPR)}, pages 991--1000, 2015.

\bibitem{wandt19}
B.~Wandt and B.~Rosenhahn.
\newblock {RepNet: Weakly Supervised Training of an Adversarial Reprojection
  Network for 3D Human Pose Estimation}.
\newblock In {\em Proceedings of the IEEE/CVF Conference on Computer Vision and
  Pattern Recognition (CVPR)}, June 2019.

\bibitem{zanfir18}
A.~Zanfir, E.~Marinoiu, and C.~Sminchisescu.
\newblock {Monocular 3D Pose and Shape Estimation of Multiple People in Natural
  Scenes - The Importance of Multiple Scene Constraints}.
\newblock In {\em Proceedings of the IEEE Conference on Computer Vision and
  Pattern Recognition (CVPR)}, June 2018.

\bibitem{zhou16}
X.~Zhou, M.~Zhu, S.~Leonardos, K.~G. Derpanis, and K.~Daniilidis.
\newblock {Sparseness Meets Deepness: 3D Human Pose Estimation from Monocular
  Video}.
\newblock {\em 2016 IEEE Conference on Computer Vision and Pattern Recognition
  (CVPR)}, pages 4966--4975, 2015.

\bibitem{zhou18}
X.~Zhou, M.~Zhu, G.~Pavlakos, S.~Leonardos, K.~G. Derpanis, and K.~Daniilidis.
\newblock {MonoCap: Monocular Human Motion Capture using a CNN Coupled with a
  Geometric Prior}.
\newblock {\em IEEE Transactions on Pattern Analysis and Machine Intelligence},
  41:901--914, 2017.

\end{thebibliography}


\end{document}